\pdfoutput=1

\documentclass[11pt]{article}

\usepackage{acl}

\usepackage{times}
\usepackage{latexsym}

\usepackage[T1]{fontenc}

\usepackage[utf8]{inputenc}

\usepackage{microtype}
\usepackage{comment}
\usepackage{inconsolata}

\usepackage{graphicx}
\usepackage{caption}

\makeatletter

\newcommand{\Rmnum}[1]{\expandafter\@slowromancap\romannumeral #1@}
\makeatother

\usepackage{booktabs}
\usepackage{colortbl}  
\usepackage{xcolor}
\usepackage{array}

\usepackage{CJKutf8}
\usepackage{amsmath}
\usepackage{subcaption}
\usepackage{multirow}
\usepackage{amssymb}
\usepackage{pifont}
\usepackage{bbding}
\usepackage{tabularx} 

\title{GLIMMER: Incorporating Graph and Lexical Features in Unsupervised Multi-Document Summarization}

\author{
 \textbf{Ran Liu\textsuperscript{1,2}},
 \textbf{Ming Liu\textsuperscript{3}\thanks{Corresponding Author. Email: m.liu@deakin.edu.au.}},
 \textbf{Min Yu\textsuperscript{1,2}\thanks{Corresponding Author. Email: yumin@iie.ac.cn.}},
 \textbf{Jianguo Jiang\textsuperscript{1,2}},
 \textbf{Gang Li\textsuperscript{3}},
 \textbf{Dan Zhang\textsuperscript{3}},
 \\
 \textbf{Jingyuan Li\textsuperscript{4}},
 \textbf{Xiang Meng\textsuperscript{1,2}},
 \textbf{Weiqing Huang\textsuperscript{1,2}}
\\
 \textsuperscript{1}Institute of Information Engineering, Chinese Academy of Sciences
\\
 \textsuperscript{2}School of Cyber Security, University of Chinese Academy of Sciences
\\
 \textsuperscript{3}School of Information Technology, Deakin University
\\
 \textsuperscript{4}Department of Computer Science, Beijing Technology and Business University
}

\begin{document}
\maketitle
\begin{abstract}
Pre-trained language models are increasingly being used in multi-document summarization tasks. However, these models need large-scale corpora for pre-training and are domain-dependent. Other non-neural unsupervised summarization approaches mostly rely on key sentence extraction, which can lead to information loss. To address these challenges, we propose a lightweight yet effective unsupervised approach called GLIMMER: a \textbf{G}raph and \textbf{L}ex\textbf{I}cal features based unsupervised \textbf{M}ulti-docu\textbf{ME}nt summa\textbf{R}ization approach. It first constructs a sentence graph from the source documents, then automatically identifies semantic clusters by mining low-level features from raw texts, thereby improving intra-cluster correlation and the fluency of generated sentences. Finally, it summarizes clusters into natural sentences. Experiments conducted on Multi-News, Multi-XScience and DUC-2004 demonstrate that our approach outperforms existing unsupervised approaches. Furthermore, it surpasses state-of-the-art pre-trained multi-document summarization models (e.g. PEGASUS and PRIMERA) under zero-shot settings in terms of ROUGE scores. Additionally, human evaluations indicate that summaries generated by GLIMMER achieve high readability and informativeness scores. Our code is available at \url{https://github.com/Oswald1997/GLIMMER}.
\end{abstract}

\section{Introduction}
Multi-document summarization (MDS) aims to produce a summary from a document set containing a series of related topics. The generated summary needs to cover all important information in the document set, while remaining fluent and concise. Compared with single-document summarization, it's more challenging because an increasing number of input documents will make the source information more redundant and scattered. It also has practical significance, for example, key information of multiple news articles can be generated efficiently. Multi-document summarization approaches can also be applied in other scenarios, such as extracting opinions from social media.

\begin{figure*}[htbp]
	\centering
	\includegraphics[width=0.95\linewidth]{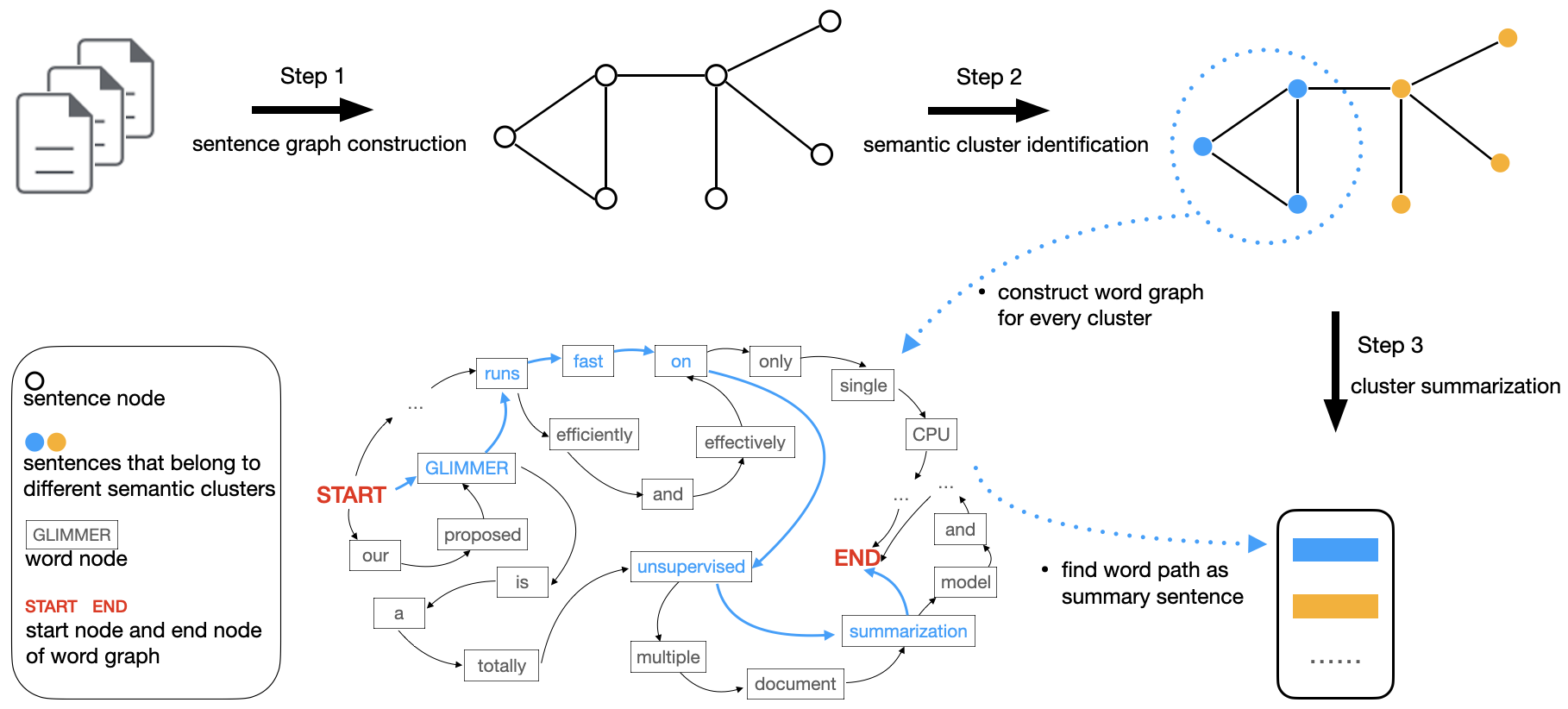}
	\caption{Illustration of GLIMMER. There are three basic steps: sentence graph construction, semantic cluster identification and cluster summarization. }
	\label{model}
\end{figure*}

As the application of MDS becomes more and more widespread, various approaches have been proposed. Non-neural approaches are primarily based on extracting key sentences \citep{erkan2004lexrank,mihalcea2004textrank,rossiello2017centroid}. These approaches assess sentence importance based on their relevance to each other or proximity to keywords, selecting sentences with high importance scores to form the summary. The main drawback of such approaches is that they retain only a subset of key sentences, which can lead to information loss and may not capture fine-grained details. 

Neural approaches can generate more abstractive text and are recently widely used in multi-document summarization. Given the structural characteristics of the multi-document input, most approaches utilize attention mechanism to build hierarchical models \citep{fabbri2019multi,mao2020multi,jin2020multi}, enabling the extraction of different-grained features and the selection of important information. Other methods employ graphs to model relationships and can leverage interaction features to enhance representation \citep{yasunaga2017graph,yin2019graph,li2020leveraging}. Despite their capability in extracting abstract features, neural models are often resource-intensive and require large parallel training datasets. Moreover, recent studies \citep{pagnoni-etal-2021-understanding-2} have shown that they also suffer from factuality problems.

Pre-trained language models use large-scale corpora to optimize objective functions, allowing them to acquire more general feature representations. This capability enables them to require only a small amount of data to fine-tune downstream tasks and achieve impressive results. Among various language models, BART \citep{lewis2019bart} and T5 \citep{raffel2020exploring} excel particularly in generation tasks. Some models leverage them as backbone and generate high-quality summaries \citep{pasunuru2021efficiently,liu2022leveraging}. However, these pre-trained models are designed for general generation tasks without specific optimizations for MDS. To address this limitation, some researchers have modified objective functions during pre-training to better align models with this task \citep{zhang2020pegasus,xiao2021primer}. Despite being the latest models, they have drawbacks such as the requirement for large-scale corpora and relatively low summarization efficiency. Moreover, the knowledge acquired during pre-training may introduce unwanted external factors that can affect the quality of summaries.

Given the challenges observed in various existing approaches, we aim to develop a general, fast, and easy-to-use MDS model that can generate high-quality summaries without annotated data or extensive computational resources. Existing supervised models rely heavily on advanced features such as word vectors and outputs from hidden layers of neural networks, overlooking important low-level features present in raw texts. Therefore, our motivation is to leverage these low-level features, such as lexical features, from raw texts to rapidly mine information and achieve unsupervised summarization. We introduce GLIMMER, illustrated in Figure~\ref{model}, which constructs a sentence graph to represent interaction relations. Unlike the graph-based methods mentioned earlier, our approach does not rely on neural layers to handle graph features. Instead, it identifies semantic clusters from the graph in an unsupervised manner. Specifically, we mine lexical features from raw texts to facilitate graph cut. Subsequently, each semantic cluster is transformed into an informative and coherent summary to generate the final output. Our primary contributions are outlined as follows:

\begin{itemize}
    \item GLIMMER is a fully unsupervised multi-document summarization approach that can run on a CPU. As an out-of-the-box solution, it does not require labeled samples, large-scale corpus or additional settings for hyperparameters, and demonstrates superiority in real-world scenarios.
    \item We utilize low-level lexical features from raw texts to automatically determine the number of semantic clusters. To our knowledge, we are the first to employ these characteristics in multi-document summarization.
    \item Experiments conducted on Multi-News, Multi-XScience, and DUC-2004 demonstrate that our approach outperforms other non-neural approaches based on automatic evaluation metrics, as well as state-of-the-art pre-trained multi-document summarization models in zero-shot scenarios. Human evaluation indicates that summaries generated by GLIMMER are more readable and factually accurate than those generated by previous non-neural approaches.
  \end{itemize}

\section{Related Work}
\label{related}

\textbf{Graph-based Text Summarization} Graphs are effective in representing relations and distances between nodes, making them well-suited for modeling text and extracting specific features. \citet{christensen2013towards} constructed an approximate discourse graph (ADG) based on discourse relations across sentences. They utilized ADG for selecting summary sentences and achieved more coherent summaries. Subsequent work has expanded upon ADG \citep{yasunaga2017graph,liu2019hierarchical} due to its ability to capture correlation features. Despite the capability of these summarization models to capture graph features, they still require training network layers to bridge the gap between features and outputs. Moreover, a large amount of supervised data is needed to train these parameters.
\\
\textbf{Topic Modelling} When identifying semantic clusters, topic models are intuitively suitable. They mine latent topics from document sets to achieve semantic clustering. Classic topic models such as Latent Dirichlet Allocation \citep{blei2003latent} and subsequent neural topic models \citep{srivastava2017autoencoding,bianchi2020cross} require the definition of the number of topics. Despite various metrics available to evaluate the quality of mined topics \citep{terragni2021octis}, determining the optimal number of topics remains challenging. BERTopic \citep{grootendorst2022bertopic} addresses this by using a hierarchical clustering algorithm to identify clusters of varying densities. However, when applied to MDS, its clustering performance is inadequate. This limitation may stem from the fact that while topic models excel at distinguishing texts with different topics, they struggle with semantic clustering where texts share similar topics. In our work, we address this challenge by constructing graphs and leveraging lexical features from raw texts, which proves effective in this scenario.
\\
\textbf{Text Compression} Some researchers focus on unsupervised sentence compression techniques. \citet{fevry2018unsupervised} add noise to original sentences and use denoising auto-encoders to recover them. \citet{ghalandari2022efficient} employ reinforcement learning to generate fluent compressed sentences of appropriate length. While these approaches are valuable, they are limited to compressing a single sentence and may fail when dealing with clusters of related sentences. To address this challenge, word graph-based methods offer a potential solution. Nodes and edges are constructed based on specific rules, and constraints are applied to identify the most suitable compression path \citep{filippova2010multi,boudin2013keyphrase,mehdad2013abstractive,shang2018unsupervised}. In our work, we integrate this concept as one module of GLIMMER and modify it to suit multi-document summarization tasks.

\section{GLIMMER Framework}
Figure~\ref{model} illustrates the framework of GLIMMER. Each input comprises multiple source documents and has been segmented into sentences. Subsequently, a sentence graph is constructed where each node represents a sentence, and each edge represents the relation between two nodes. Graph cut is then applied to the affinity matrix to identify subgraphs based on lexical features, with each subgraph representing a semantic cluster. Next, we construct a directed word graph for each cluster. By selecting the most suitable path within each graph, we generate informative and fluent summary sentences. The three main modules of GLIMMER are detailed below.

\subsection{Sentence Graph Construction}
\label{Construct Sentence Graph}
Given that ADG can effectively describe discourse relations and reflect the degree of correlation, it aligns well with our task. The original implementation of ADG relies on specific indicators such as co-reference and discourse cues, it also needs statistics from large corpus. To streamline the process and enhance the effectiveness of subsequent graph cut operations, we modify ADG and construct sentence graph based on the following four indicators:
\\
\textbf{Deverbal Noun Reference}
A verb is associated with describing an event, and in subsequent sentence, the nominalization form of this verb is usually used to refer to the event. This type of expression can serves as an indicator to connect related sentences. Specifically, we extract verbs from a sentence,\footnote{We use spaCy for implementation. \url{https://spacy.io}.} filtering out non-notional verbs such as \textit{was} or \textit{had}. We then use WordNet to obtain all relevant nominalization forms of verbs.\footnote{We use NLTK for implementation. \url{https://www.nltk.org}.} Next, we identify the most similar words for these nouns using word vectors, resulting in a list of nouns to be matched. When processing subsequent sentences, we compare their nouns to the list generated earlier. A match indicates that these two sentences contain a deverbal noun reference relation.
\\
\textbf{Conjunctions}
A conjunction is used to join two sentences, indicating a relationship between them. This relationship can be either coordinative or adversative, suggesting that the sentences are related to the same topic. In our implementation, we identify 39 conjunctions. If a subsequent sentence begins with one of these conjunctions, then the two sentences are considered to match successfully.
\\
\textbf{Entity Consistency}
If two sentences have the same entity, it means they likely refer to the same event, indicating semantic correlation. To implement this, we extract named entities from two sentences. If they contain the same entity with the same entity type, we consider it a successful match.
\\
\textbf{Semantic Similarity}
This is a straightforward indicator. We calculate sentence vectors and use cosine value to measure the similarity between two sentences. If cosine similarity exceeds a certain threshold, we consider the match successful.

Based on the aforementioned four indicators, we construct graph $(V, E)$, where each node $v_i \in V$ represents a single sentence, and each edge $e_{i,j} \in E$ indicates connection status between node $v_i$ and node $v_j$. We set $e_{i,j} = 1$ if at least one of the four indicators is satisfied, or $e_{i,j} = 0$ if none conditions is met. Note that we only use deverbal noun reference and conjunctions indicators for adjacent sentences, as it is not meaningful when sentences are far apart in text. We don't set edge weights like \citet{christensen2013towards}, because although weighted ADG can aid in path selection and contribute to get coherent summary sentences, it shows limited improvement in subsequent graph cut operations.

\subsection{Semantic Cluster Identification}
\label{Find Semantic Clusters}
After finishing the construction of the sentence graph, we obtain an $n \ast n$ adjacency matrix, where $n$ is equal to the number of input sentences. This matrix can be considered as an affinity matrix and is subjected to graph cut to identify normalized graph cuts, with each resulting subgraph representing a semantic cluster. As discussed in §\ref{related}, determining the appropriate number of semantic clusters is challenging. Therefore, in this section, we propose two methods to address this issue, called TTR-based method and distance-based method, respectively. Then graph cut can be conducted.
\subsubsection{Clustering Number Determination}
\textbf{TTR-based Method}
Lexical features can reflect semantic information to some extent. Among different features, type-token ratio (TTR) is one of the most commonly used feature and it can measure lexical richness. The formula is as follows where $T$ represents the number of unique words and $N$ represents the total number of words.
\begin{eqnarray}
\label{ttr:true}
  TTR = \frac{T}{N} 
\end{eqnarray}
Intuitively, TTR is positively correlated with semantic richness. This is because richer semantic information often leads to a richer vocabulary, whereas semantically poor expressions tend to contain more repetitive words. Similarly, a larger number of semantic clusters tends to imply richer semantic information. As a consequence, TTR of the input text is also positively correlated with the number of semantic clusters or subgraphs. A simple formula to estimate number of clusters can be expressed as follows:
\begin{eqnarray}
\label{ttr:original}
  n_{cluster} = \lfloor n_{sent} \ast TTR_{input}\rfloor 
\end{eqnarray}
The formula calculates the number of semantic clusters based on the product of the whole input's TTR value and the number of sentences in the input. In an extreme case where TTR equals one, each sentence in the input exhibits high semantic richness and is distinct from others, resulting in each sentence forming a semantic cluster independently. However, this formula assumes equal contribution from all input sentences when calculating the number of clusters, thereby overlooking the varying influences of different sentences on semantic richness. Specifically, sentences with higher TTR values contain richer information, leading to a greater number of clusters, and vice versa. To address this, we differentiate between high-TTR and low-TTR sentences. We refer to \citet{mckee2000measuring}, who proposed a method to illustrate the relation between TTR and sample size, the formula is as follows:
\begin{eqnarray}
\label{ttr:estimate}
  TTR = \frac{D}{N}  [(1+2\frac{N}{D})^{\frac{1}{2}} - 1]
\end{eqnarray}
In this formula, $D$ is used as a parameter of calculating lexical richness. It is estimated from original input, please refer to Appendix D for detailed information on the estimation process. Once $D$ is determined, we apply Formula~(\ref{ttr:estimate}) to each sentence of the input to obtain the estimated TTR value for each sentence. Additionally, we calculate the true TTR value for each sentence using Formula~(\ref{ttr:true}).
\begin{eqnarray}
  \label{ttr:low}
  \frac{TTR_{true}}{TTR_{esti}} < 1 - \sigma 
\end{eqnarray}
\begin{eqnarray}
  \label{ttr:high}
  \frac{TTR_{true}}{TTR_{esti}} \geqslant  1 + \sigma 
\end{eqnarray}
If a sentence from the input satisfies Formula~(\ref{ttr:low}), it is considered a low-TTR sentence, otherwise, it is classified as a high-TTR sentence. Furthermore, we extend Formula~(\ref{ttr:original}) to Formula~(\ref{ttr:final}), where $n_{low}$ and $n_{high}$ represent the numbers of low-TTR and high-TTR sentences, respectively. $\beta$ is an influence factor that controls the degree to which different sentences affect the semantic richness of the overall input.
\begin{eqnarray}
  \label{ttr:final}
  \begin{split}
  n_{cluster} = \lfloor [ n_{sent} - \beta (n_{low} - n_{high})] \\ \ast TTR_{input}\rfloor 
  \end{split}
\end{eqnarray}
\\
\textbf{Distance-based Method}
This method is much simpler and more intuitive than TTR-based method, it utilizes the adjacency matrix calculated from §\ref{Construct Sentence Graph}. The core idea is similar to silhouette coefficient, aiming to minimize the average distance between nodes within the same cluster and maximize the average distance between nodes in different clusters. To implement this method, we vary the number of clusters and apply spectral clustering algorithm to obtain different graph cut results. We then use Floyd-Warshall algorithm to calculate distances between nodes and determine the optimal number of clusters that satisfy the aforementioned requirements.
\subsubsection{Graph Cut}
After obtaining the optimal number of semantic clusters using either the TTR-based or distance-based method, we perform graph cut on the affinity matrix to find normalized graph cuts. Since our affinity matrix and adjacency matrix are identical, there is no need to construct a similarity graph. We calculate Laplacian matrix $L$ using Formula~(\ref{laplacian}), where $W$ is adjacency matrix and $D$ is degree matrix.
\begin{eqnarray}
  \label{laplacian}
  L = D - W 
\end{eqnarray}

Next, we calculate first $k$ eigenvectors $u_1,u_2,...,u_k$ of $L$, where $k$ is the number of clusters that we have determined. We then construct eigenmatrix $U \in \mathbb{R}^{n \ast k}$, which consists of eigenvectors $u_1,u_2,...,u_k$ as columns. Finally, we apply k-means algorithm to partition the $n$ sentences into $k$ clusters, resulting in the formation of semantic clusters.

\subsection{Semantic Cluster Summarization}
\label{Summarize Semantic Clusters}
We perform multi-sentence summarization on each semantic cluster to generate a concise summary which is also informative and fluent. 

The first step is to construct a word graph for each cluster. Figure~\ref{model} contains an example of word graph. Each graph has a start node and an end node, and each sentence in the cluster is connected between these two special nodes sequentially and separately. The edges in the graph follow the direction of natural language, with each node representing a single word. To illustrate interactions among sentences within the same cluster, words with the same lowercase form and part of speech are mapped to identical nodes. As a result, the number of paths between the start and end nodes becomes significantly larger than the number of sentences in the cluster. It's important to note that no two words within a single sentence will be mapped to the same node. In cases where there are multiple choices for mapping subsequent words to the graph, we examine the context of the nodes to select the node with the most context coincidence or the highest mapping frequency.

For edge weights, we refer to \citet{filippova2010multi}:
\begin{eqnarray}
  w(e_{i, j})=\frac{f(i) + f(j)}{\sum_{s \in S} D_s(i,j)^{-1} } \cdot \frac{1}{f(i) \ast  f(j)} 
\end{eqnarray}
$f(i)$ represents frequency or mapped times of node $i$. $D_s(i,j)^{-1}$ denotes the inverse value of positional distance between node $i$ and $j$ in sentence $s$, which is meaningful only when node $i$ precedes node $j$. The left term indicates that when selecting the shortest path, preference is given to edges where nodes have low frequencies but strong connections. The right term suggests that important nodes are more likely to be selected. In this way, the selected path contains both salient words and specific expressions, making the summary of a cluster informative.

Once edge weights are determined, we select the shortest path from start node to end node, which has the smallest sum of edge weights. Additional, to further improve fluency, we divide the sum of edge weights by the sum of n-gram probabilities for each path. In this way, we reselect path with the lowest score. After summarizing of all semantic clusters, we obtain a final summary of the input documents set.

\section{Experiments}
\subsection{Datasets}
We use \textbf{Multi-News} \citep{fabbri2019multi}, \textbf{Multi-XScience} \citep{lu2020multi} and \textbf{DUC-2004} \citep{over2004introduction} as datasets, all of which are commonly-used MDS datasets. We have verified that these datasets do not contain sensitive data in terms of privacy and security. As an unsupervised approach, we only use test set of the above datasets, with document set sizes of 5622, 5093 and 50 respectively. More descriptions of datasets can be found in Appendix A.1.

\subsection{Baselines}
Our baselines are \textbf{First-n}, \textbf{LexRank} \citep{erkan2004lexrank}, \textbf{Centroid} \citep{rossiello2017centroid}, \textbf{Summpip} \citep{zhao2020summpip}, \textbf{PEGASUS} \citep{zhang2020pegasus}, \textbf{PRIMERA} \citep{xiao2021primer}. Please refer to Appendix A.2 for details.

Inspired by determining clustering number based on eigengap, we replace our TTR-based and distance-based methods with eigengap-based method and obtain a new baseline called \textbf{GLIMMER-Eigengap}.

\subsection{Experiment Design}
Since Multi-News dataset provided by \citet{fabbri2019multi} has been tokenized, we tokenize Multi-XScience and DUC-2004 as well to standardize the data format and facilitate performance comparison across different approaches.

Following \citet{xiao2021primer}, we control output lengths of baselines to ensure fair comparison. As the performance of each baseline model varies significantly across different datasets depending on hyperparameters, we adjusted these hyperparameters separately for each dataset to ensure that baselines performs well across different datasets. For more details, please refer to Appendix A.

For GLIMMER, we set thresholds for identifying similar words and matching similar sentences in §\ref{Construct Sentence Graph} to $0.65$ and $0.98$ respectively. Additionally, We set $\sigma = 0.05$ and $\beta = 4$ for TTR-based method.

\begin{table*}[ht]
    \caption{\label{result-main}
    Rouge scores of different models on Multi-News, Multi-XScience and DUC-2004. GLIMMER-TTR and GLIMMER-Distance indicates our approach utilizes TTR-based and distance-based method described in §\ref{Find Semantic Clusters} respectively. Best results under each category have been \textbf{bolded} (statistical significance with p-value $<$ 0.05). Results of PEGASUS on all datasets, PRIMERA on Multi-XScience and DUC-2004 are from \citet{xiao2021primer}.
    }
  \centering
  \begin{tabular}{lccccccccc}
  \toprule
  & \multicolumn{3}{c}{\textbf{Multi-News}} & \multicolumn{3}{c}{\textbf{Multi-XScience}} & \multicolumn{3}{c}{\textbf{DUC-2004}}\\
model & R-1 & R-2 & R-L & R-1 & R-2 & R-L & R-1 & R-2 & R-L \\
\midrule
First & 38.16 & 11.14 & 18.67 & 23.52 & 3.29 & 13.27 & 26.17 & 6.38 & 14.34 \\
\midrule
LexRank & 40.59 & 11.99 & 19.57 & 30.00 & 4.80 & \textbf{17.42} & 34.84 & 7.50 & 19.77 \\
Centroid & 41.65 & 12.40 & 19.85 & 31.18 & 4.65 & 17.05 & 37.17 & 8.08 & 19.52 \\
Summpip & 40.99 & 11.91 & 19.02 & 29.24 & 4.07 & 16.58 & \textbf{37.35} & \textbf{8.57} & 19.77 \\
\midrule
PEGASUS & 32.10 & 10.10 & 16.70 & 27.60 & 4.60 & 15.30 & 32.70 & 7.40 & 17.60 \\
PRIMERA & 32.93 & 9.12 & 17.83 & 29.10 & 4.60 & 15.70 & 35.10 & 7.20 & 17.90 \\
\midrule
GLIMMER-TTR & \textbf{43.08} & \textbf{13.87} & \textbf{21.05} & \textbf{31.79} & \textbf{4.81} & 16.71 & 34.90 & 6.90 & 18.45 \\
GLIMMER-Distance & 42.44 & 13.26 & 21.00 & 31.56 & 4.59 & 16.46 & 35.27 & 7.66 & \textbf{20.31} \\
GLIMMER-Eigengap & 28.83 & 9.05 & 16.50 & 30.38 & 4.44 & 17.14 & 28.74 & 6.64 & 17.86 \\
  \bottomrule
  \end{tabular}

\end{table*}

\begin{table}[ht]
    \caption{\label{fine-tune}
  Zero and few-shot results of PRIMERA on Multi-News.
  }
  \centering
  \begin{tabular}{lccc}
  \toprule
  model & R-1 & R-2& R-L\\
  \midrule
  PRIMERA (0) & 32.93 & 9.12 & 17.83 \\
  PRIMERA (10) & 40.86 & 12.31 & 21.09 \\
  PRIMERA (100) & 43.02 & 13.58 & 22.10 \\
  \bottomrule
  \end{tabular}
  \end{table}

  \begin{table}[ht]
    \caption{\label{bertscore}
    BERTScore results of GLIMMER and baslines on Multi-News. Baselines are non-neural SOTA Summpip and neural SOTA PRIMERA. Both zero-shot and fully fine-tuned PRIMERA are included.
    }
    \centering
    \begin{tabular}{lccc}
    \toprule
    model & P & R& F1\\
    \midrule
    Summpip &60.45 & 58.38& 59.33\\
    PRIMERA (0) & 58.86 & 52.92 & 55.62 \\
    PRIMERA (full) & \textbf{60.90} & 58.01 & 59.36 \\
    GLIMMER-TTR & 59.79 & \textbf{60.43} & \textbf{59.96} \\
    GLIMMER-Distance & 60.50 & 59.21 & 59.63 \\
    \bottomrule
    \end{tabular}
    \end{table}

\subsection{Automatic Evaluation}
We use ROUGE scores to automatically evaluate summarization performance. Specifically, we use $f_1$ of ROUGE-1, ROUGE-2 and ROUGE-L,\footnote{We use ROUGE-1.5.5 implemented by \url{https://github.com/li-plus/rouge-metric}.} which take into account completeness, readability and order of summaries.
  
Table~\ref{result-main} presents results of GLIMMER and baselines. Our approach achieves the best results on two-thirds of metrics, indicating improvements in coverage and word sequence quality in multi-document summarization. GLIMMER performs particularly well on two datasets, including Multi-XScience, which is more challenging due to its higher abstractiveness. This demonstrates that GLIMMER is not only suitable for news articles but can also be applied to other fields such as technical articles. Furthermore, results of GLIMMER surpass non-neural unsupervised approaches and large-scale pre-trained models. However, GLIMMER-Eigengap performs poorly on Multi-News, suggesting that this traditional method for determining clustering numbers does not generalize well across different tasks.

However, we observed that pre-trained models like PEGASUS and PRIMERA do not outperform traditional non-neural network approaches, despite being considered SOTA summarization models. This could be attributed to their pre-training on large corpora, which may have distributions different from that of the test sets. Table~\ref{fine-tune} indicates that only when PRIMERA is fine-tuned using supervised samples does its excellent feature extraction ability become apparent. However, even under few-shot scenarios, GLIMMER's ROUGE scores remain competitive with those of fine-tuned models.

In addition, GLIMMER also achieves highest scores on BERTScore \citep{bertscore},\footnote{In implementation, \textit{deberta-xlarge-mnli} is used to represent embeddings.} even outperforming fully fine-tuned PRIMERA (Table~\ref{bertscore}). We also use a neural evaluation framework called UniEval \citep{zhong2022towards} to automatically assess the readability of summaries. While not as accurate as human evaluation, we consider this an additional experiment, and the results can be found in Appendix C.

\subsection{Human Evaluation}
Automatic evaluation alone does not provide a comprehensive assessment of summarization quality, as readability and informativeness are also important factors to consider. We use fluency, coherence, and referential clarity as indicators to evaluate readability. Together with informativeness, these four indicators are used as the basis for human evaluation on 50 random samples. For more details on the human evaluation process, please refer to Appendix B.1.

\begin{table*}[ht]
    \caption{\label{human}
    Results of Human Evaluation.
    }
  \centering
  \begin{tabular}{lcccc}
  \toprule
  model & fluency & coherence & referential clarity & informativeness\\
  \midrule
  Summpip & 2.95 & 1.91 & 2.08 & 2.50 \\
  PRIMERA & \textbf{3.68} & \textbf{2.76} & \textbf{2.95} & 2.57 \\
  GLIMMER-TTR & 3.54 & 2.50 & 2.83 & \textbf{3.67} \\
  GLIMMER-Distance & 3.15 & 2.42 & 2.52 & 3.10 \\
  \bottomrule
  \end{tabular}
  \end{table*}

  Table~\ref{human} presents the average evaluation results from three annotators for different models. Summpip shows the worst overall performance, while our approach demonstrates significant improvement over previous non-neural methods. PRIMERA, being a large language model, achieves the highest readability scores, which is expected given its extensive language knowledge learned during pre-training and its abstractive nature. GLIMMER-TTR closely follows PRIMERA, also obtaining relatively high readability scores. Additionally, its informativeness score significantly exceeds that of PRIMERA, suggesting that the summaries generated by our model not only contain more information but also match the state-of-the-art neural model in terms of readability. For more details regarding agreement analysis and score distributions, please refer to the Appendix B.

\section{Analysis}
\label{Analysis}
\subsection{Ablation}
The basic structure of GLIMMER consists of three main components: sentence graph construction, semantic clustering, and cluster summarization. In the second module, one of our main contributions is devising methods to automatically determine the number of clusters. In our ablation study, similar to most clustering algorithms, we set the number of clusters to a fixed value, which is a commonly used approach. Then, in the third module, we select the shortest path based solely on path weights, disregarding fluency factor.

Results of ablation study on Multi-News are presented in Table~\ref{ablation}. As predicted, base model performs the worst. It neither utilizes strategies to identify semantic clusters nor considers fluency when selecting paths, resulting in relatively low performance. Both our TTR-based and distance-based methods show improvements in results. Specifically, TTR-based method performs better than distance-based method. This is because TTR-based method directly extracts fundamental features from raw texts, whereas distance-based method relies on sentence-correlation features, which are more complex and may conflict with raw text.

The addition of fluency constraint also contributes to the improvement of results. Furthermore, we observed that if neither TTR-based nor distance-based method is used, incorporating the fluency factor leads to a significant improvement. However, the improvement is less pronounced when TTR or distance is already employed. This observation highlights the effectiveness of TTR-based and distance-based method from another perspective. A cluster of closely related sentences makes it easier to summarize fluently, thereby reducing the importance of the fluency factor.

\begin{table}[ht]
    \caption{\label{ablation}
    Ablation study on Multi-News, base means the basic structure of GLIMMER with clusters number a fixed value of 9 and ignoring fluency when selecting path.
    }
  \centering
  \begin{tabular}{lccc}
  \toprule
  model & R-1 & R-2& R-L\\
  \midrule
  base & 38.23 & 10.93 & 18.38 \\
  w/ fluency & 41.37 & 12.31 & 19.26 \\
  w/ distance & 40.88 & 12.31 & 20.50 \\
  w/ distance, fluency & 42.44 & 13.26 & 21.00 \\
  w/ TTR & 42.05 & 12.96 & 20.61 \\
  w/ TTR, fluency & 43.08 & 13.87 & 21.05 \\
  \bottomrule
  \end{tabular}
  \end{table}

  \begin{table}[ht]
    \caption{\label{neural-compression}
    Results of adopting neural summarization models.
    }
    \centering
    \begin{tabular}{lccc}
    \toprule
    model & R-1 & R-2& R-L\\
    \midrule
    GLIMMER-TTR & \textbf{43.08} & \textbf{13.87} & \textbf{21.05} \\
    newsroom-L11 & 33.11 & 8.31 & 16.51 \\
    newsroom-P75 & 41.01 & 11.31 & 18.98 \\
    multi-news-P40 & 35.76 & 9.58 & 16.91 \\
    multi-news-Gamma & 33.70 & 8.87 & 16.17 \\
    \bottomrule
    \end{tabular}
    \end{table}

  \subsection{Cluster Summarization by Neural Models}
  As described in §\ref{Summarize Semantic Clusters}, we generate summary sentences in a non-neural way. In this section, we explore the effectiveness of replacing our approach with neural approaches. Specifically, we adopt the approach proposed by \citet{ghalandari2022efficient} and directly use their trained models (newsroom-L11 and newsroom-P75) that were trained on Newsroom \citep{grusky2018newsroom}. Additionally, we trained two new models on Multi-News using their training strategies. We integrate these neural models into step 3 of GLIMMER while keeping the other steps unchanged. Please refer to Table~\ref{neural-compression} for ROUGE scores obtained on Multi-News and Appendix G for details about these neural models. 
  
We find that using neural models for generation does not lead to improvements in ROUGE scores. Neural models require large corpora and substantial computing resources for training, which is not aligned with our goals. Moreover, the effectiveness of text generation is highly dependent on the parameter settings during training, resulting in limited generalization capabilities.

\subsection{Comparison with ChatGPT}
  ChatGPT demonstrates remarkable dialogue abilities and can be applied to various tasks, including summarization. The fundamental principles of ChatGPT align with InstructGPT \citep{ouyang2022training}, utilizing reinforcement learning from human feedback (RLHF) to optimize the language model. ChatGPT is not only pre-trained on a large-scale corpus but also fine-tuned using supervised data, with significant human effort dedicated to training a reward model. Given that ChatGPT is regularly updated and publicly available, it is highly likely that ChatGPT has been exposed to some of datasets used in our experiments. Nevertheless, we consider ChatGPT to represent an upper bound for GLIMMER.
  
  We use GPT-3.5-based ChatGPT for experiments, specifically, we employ gpt-3.5-turbo due to its capability and cost-effectiveness. Additionally, we utilize text-davinci-003 for further comparison. Due to API limitations, we tested only the first 50 samples from each dataset. Prompt and more detailed information can be found in Appendix E.1.

  Comparison results with ChatGPT based on ROUGE scores are presented in Table~\ref{rouge-gpt}. Despite facing a powerful language model, our approach still performs well, particularly on Multi-News and Multi-XScience. However, regarding DUC-2004, which was released earlier, we hypothesize that ChatGPT may have already been trained on these texts, resulting in substantially better performance compared to other datasets.
  
  Human evaluation is conducted to assess the readability of summaries generated by ChatGPT. Our findings indicate that both gpt-3.5-turbo and text-davinci-003 exhibit higher fluency, coherence, and referential clarity. However, a notable disadvantage is the potential for ChatGPT to produce unfaithful outputs, with a higher likelihood observed when using the text-davinci-003 model. It is evident that each model mentioned exhibits trade-offs among different indicators, and it remains challenging to identify a summary model that excels in all aspects. For detailed human evaluation results and examples of hallucinated outputs, please refer to Appendix E.2.

  \begin{table*}[ht]
    \caption{\label{rouge-gpt}
    Comparison results with ChatGPT.
    }
    \centering
    \begin{tabular}{lccccccccc}
    \toprule
    & \multicolumn{3}{c}{\textbf{Multi-News}} & \multicolumn{3}{c}{\textbf{Multi-XScience}} & \multicolumn{3}{c}{\textbf{DUC-2004}}\\
  model & R-1 & R-2 & R-L & R-1 & R-2 & R-L & R-1 & R-2 & R-L \\
  \midrule
  gpt-3.5-turbo & 42.03 & 11.71 & 20.73 & \textbf{31.98} & 3.61 & \textbf{16.82} & 40.86 & 10.34 & 21.71 \\
  text-davinci-003 & 42.00 & 12.66 & 21.28 & 30.67 & 3.83 & 15.78 & \textbf{41.13} & \textbf{11.04} & \textbf{22.72} \\
  GLIMMER-TTR & \textbf{43.97} & \textbf{14.86} & \textbf{21.47} & 31.97 & \textbf{4.24} & 16.05 & 35.27 & 7.66 & 20.31 \\
  GLIMMER-Distance & 43.23 & 14.08 & 21.42 & 30.60 & 3.52 & 15.86 & 34.90 & 6.90 & 18.45 \\
    \bottomrule
    \end{tabular}
  \end{table*}

\subsection{Clustering Visualization}
    We visualize the clustering results of TTR-based and distance-based methods. We compare them with the base model which set the number of clusters to 9 because this setting achieves relatively good ROUGE scores on Multi-News.
    
    For TTR-based method, we reduce the dimension of eigenmatrix using PCA and UMAP. Since different cluster numbers correspond to different eigenmatrices, position distributions of nodes after dimensionality reduction varies. Please refer to Appendix F.1 for visualization results on Multi-News. Each node represents a sentence, and nodes of the same color indicate that they belong to the same semantic cluster. It is evident that TTR-based method produces better clustering results, while the base model appears to exhibit more random clustering.
    
    For distance-based method, we visualize the adjacency matrix calculated in §\ref{Construct Sentence Graph}, as it determines the number of clusters based on the distances between nodes in the adjacency matrix. This visualization provides a more intuitive representation. Results of graph cut based on distance, shown in Appendix F.2, outperform the base model, indicating the effectiveness of this method.

\subsection{Case Study}
    \label{sec:case}
\begin{figure}[htbp]
        \centering
        \includegraphics[width=0.98\linewidth]{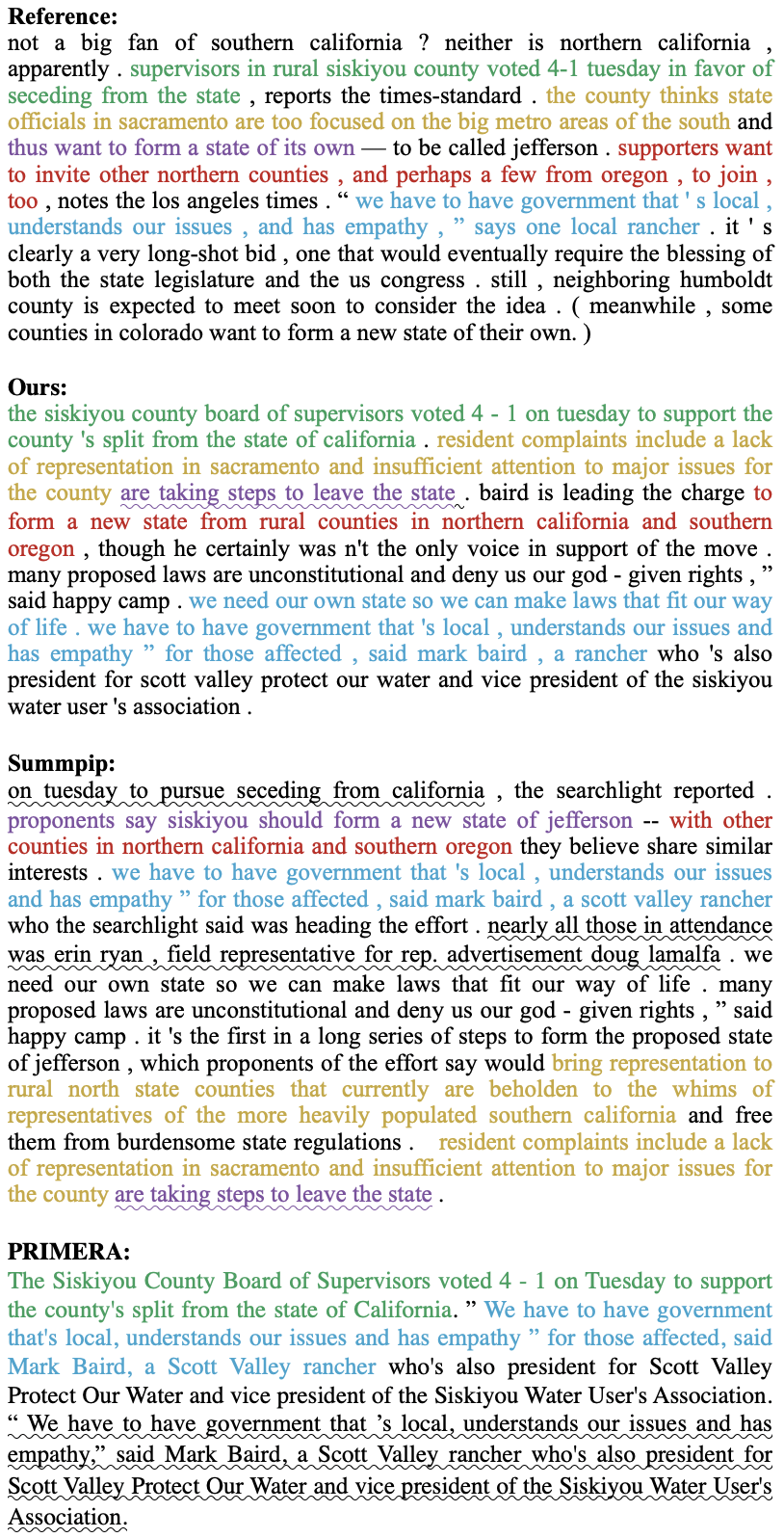}
        \caption{Comparisons between reference and generated summaries.}
        \label{case}
    \end{figure}

    Figure~\ref{case} shows comparisons between reference and generated summaries. Parts with the same color indicate similar meanings, while text that is underlined represents grammatical errors or redundancy. 
    
    From these examples, we can conclude that compared to Summpip, the summary generated by our approach is more comprehensive and contains less irrelevant information. Additionally, our approach exhibits fewer grammatical errors. Regarding PRIMERA, it tends to produce repeated sentences, resulting in high redundancy.
    
\section{Conclusion}
    Our proposed GLIMMER is effective and efficient for unsupervised multi-document summarization, requiring no additional data before use. By leveraging basic features of raw texts and mining semantic clusters, GLIMMER generates high-quality summaries. Results demonstrate that GLIMMER outperforms current unsupervised approaches and even state-of-the-art pre-trained models under zero-shot settings. Moreover, our approach shows competitiveness with ChatGPT on certain datasets. In future work, we plan to incorporate external knowledge to generate more abstractive summaries.

\section{Ethics Statement}
    All data and codes used in this paper comply with the license for use. GLIMMER poses minimal data storage and leakage risks because it has no trainable parameters, and inferring original texts from the generated summaries is almost impossible. However, there remains a small probability of generating biased or toxic texts during the path selection in word graphs.
    
    Data collection approval was received from an ethics review board. The remuneration paid to the annotators exceeds the average salary level in the area where they are located.

\section*{Acknowledgments}
    This work is supported by Youth Innovation Promotion Association CAS (No.2021155).

\begin{CJK}{UTF8}{gbsn}
\bibliography{anthology,custom}
\end{CJK}

\newpage
\appendix

\section{Experiments Details}
\label{sec:experiments}

\subsection{Datasets Details}
\label{dataset_detail}
\textbf{Multi-News}
\citep{fabbri2019multi} A widely used dataset for multi-document summarization, featuring high-quality news article summaries professionally written by editors.
\\
\textbf{Multi-XScience} 
\citep{lu2020multi} A more challenging dataset by focusing on scientific articles. Each source text comprises an article's abstract along with abstracts from its referenced articles. The summary is derived from the related work section of the main article.
\\
\textbf{DUC-2004} 
A classic multi-document summarization dataset. Each documents set consists of 10 news articles and 4 reference summaries.

\subsection{Baselines Details}
\label{baseline_detail}
\textbf{First-n}
We extract the first $n$ sentences from each article in a document set and combine them to form a summary of the set. This method works because summaries often contain key information found at the beginning of articles.
\\
\textbf{LexRank}
\citep{erkan2004lexrank} A graph-based approach where sentences are represented as nodes, and edges represent similarities between sentences. It selects summary sentences based on their relation to other sentences in the document.
\\
\textbf{Centroid}
\citep{rossiello2017centroid} A method based on identifying the most relevant words and calculating a centroid embedding. Sentences are scored based on their similarity to this centroid.
\\
\textbf{Summpip}
\citep{zhao2020summpip} An unsupervised pipeline designed for summarizing multiple documents. It performs well on various tasks including opinion summarization.
\\
\textbf{PEGASUS}
\citep{zhang2020pegasus} A Transformer-based model designed for abstractive summarization. It is pre-trained on a large corpus using self-supervised objectives. We use PEGASUS as a baseline under zero-shot settings.
\\
\textbf{PRIMERA}
\citep{xiao2021primer} A pre-trained model specialized in multi-document representation, excelling in connecting and aggregating information across documents. We deploy PRIMERA under zero-shot settings, as well as few-shot settings using 10 and 100 examples for better comparative analysis.

\subsection{Baselines Settings}
\textbf{First} We extract the first 2, 1, 1 sentence from each article in a document set for Multi-News, Multi-XScience, DUC-2004, respectively.
\\
\textbf{LexRank} Number of key sentences is set to 6, 3, 4 for Multi-News, Multi-XScience, DUC-2004, respectively.
\\
\textbf{Centroid} We use glove-wiki-gigaword-100\footnote{\url{https://github.com/RaRe-Technologies/gensim-data}} as word vectors. Similar to LexRank, we set the number of key sentences to 6, 3, 4 for three datasets.
\\
\textbf{Summpip} We set the cluster number to 9, 7, 5 for three datasets. The minimal length of each summary sentence is no less than 6 words.
\\
\textbf{PEGASUS} For few-shot PRIMERA with 10 and 100 examples, we train it for 200 and 100 epoches respectively. We set learning rate to $3e-5$, and a warm-up strategy is adopted. Validation check will be conducted every 5 epochs.

\subsection{Length Control}
Following \citet{fabbri2019multi}, we truncate the input of Multi-News and DUC-2004 to 500 tokens, which is a commonly used pre-processing step. For each sample with $S$ source input documents, we extract the first $N/S$ tokens from each source document, where $N$ is the total desired input length. Since some source documents may be shorter, we iteratively determine the number of tokens to extract from each document until the desired length is reached. Regarding Multi-XScience, we do not truncate it because its average input length is not significantly different from 500 tokens. Additionally, we truncate DUC-2004 to different lengths (500, 1000, 1500) and experiment with GLIMMER. The results are presented in Table \ref{truncate}. As noted in \citep{fabbri2019multi}, increasing the input length does not significantly improve the results.

For the summary length, to ensure fair comparison among baselines, we follow \citet{xiao2021primer} and set a uniform output length. Specifically, we use lengths of 256, 128, and 128 for Multi-News, Multi-XScience, and DUC-2004, respectively. This is implemented by adjusting hyperparameters or truncating outputs accordingly.

\begin{table}[ht]
    \caption{\label{truncate}
    Experiments with different truncated length of DUC-2004.
    }
  \centering
  \begin{tabular}{llc}
  \toprule
  \multicolumn{2}{c}{Length} & GLIMMER-TTR \\
  \midrule
  \multirow{3}*{500}&R-1&34.90 \\
  ~&R-2&6.90 \\
  ~&R-L&18.45 \\
  \midrule
  \multirow{3}*{1000}&R-1&34.36 \\
  ~&R-2&6.56 \\
  ~&R-L&17.70 \\
  \midrule
  \multirow{3}*{1500}&R-1&34.87 \\
  ~&R-2&6.92 \\
  ~&R-L&18.01 \\
  \bottomrule
  \end{tabular}
\end{table}

\subsection{Other Details}
The experiments regarding GLIMMER are conducted on a machine equipped with Intel(R) Core(TM) i7-6700 CPU @ 3.40GHz, without utilizing any GPUs. The experiments regarding neural network models are conducted on NVIDIA GeForce RTX 2080 Ti.

Regarding runtime, GLIMMER is several times or even tens of times faster than neural network models, including traditional neural networks, pre-trained language models, and large language models. For example, we estimated the time required for LLMs on the Multi-News test set. The local \texttt{vicuna-13b-v1.5} \citep{zheng2023judging} and \texttt{chatglm3-6b} \citep{du-etal-2022-glm-2} took over 30 and 10 hours respectively on an NVIDIA A100 40GB PCIe GPU, while online LLMs took even longer due to request limitations. In contrast, GLIMMER can reduce the time to 2 hours (on an Intel Core i7-6700 CPU, due to the low CPU usage, parallel computing can be utilized).

\subsection{Fully Supervised Results}
\label{fully}
Apart from few-shot fine-tuned results, fully supervised fine-tuned results are also crucial for understanding the extent to which our approach can be improved. Table~\ref{fully-result} displays the fully supervised results of PEGASUS, BART-Long-Graph, and PRIMERA on Multi-News. Although GLIMMER outperforms existing unsupervised approaches, it still lags behind mainstream fully supervised models.

\begin{table}[ht]
    \caption{\label{fully-result}
    Fully supervised fine-tuned results on Multi-News. C4 and HugeNews are pre-training corpora.
    }
  \centering
  \begin{tabular}{lccc}
  \toprule
  model & R-1 & R-2& R-L\\
  \midrule
  GLIMMER-TTR & 43.08 & 13.87 & 21.05 \\
  \midrule
  PEGASUS$\rm_{C4}$ & 46.74 & 17.95 & 24.26 \\
  PEGASUS$\rm_{HugeNews}$ & 47.52 & 18.72 & 24.91 \\
  BART-Long-Graph & 49.24 & 18.99 & 23.97 \\
  PRIMERA & 49.90 & 21.10 & 25.90 \\
  \bottomrule
  \end{tabular}
  \end{table}

  \section{Details of Human Evaluation}
  \label{sec:human}
  
  \subsection{Task Descriptions}
  \label{task_descriptions}
\begin{figure*}[ht]
    \centering
    \includegraphics[width=0.7\linewidth]{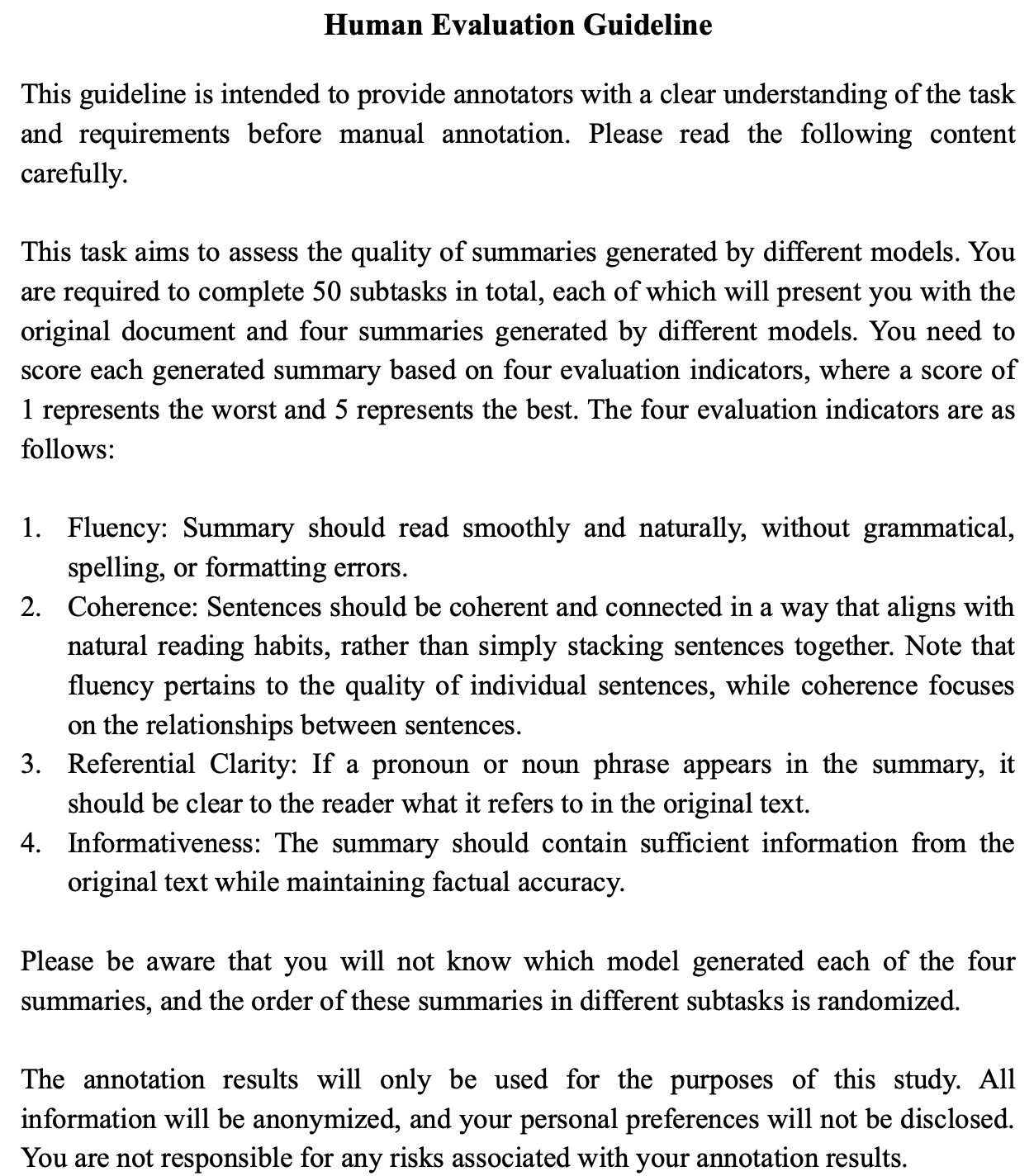}
    \caption{Human evaluation guideline}
    \label{guideline}
  \end{figure*}

  Three annotators\footnote{Graduate students, two of whom are non-native English speakers and one is a native English speaker.} score summaries independently. Each annotator need to complete 50 subtasks, with each subtask consisting of a source text and four summaries generated by Summpip, PRIMERA (zero-shot), GLIMMER-TTR, and GLIMMER-Distance, respectively. All summaries are lowercased and tokenized, which makes it impossible to use capitalization and other textual features to identify which summaries are likely generated by the same model. We have developed guidelines for the annotators, as shown in Figure~\ref{guideline}.

  \subsection{Score Distributions}
  \label{scores_distributions}
\begin{figure*}[htbp]
    \centering
    \includegraphics[width=0.95\linewidth]{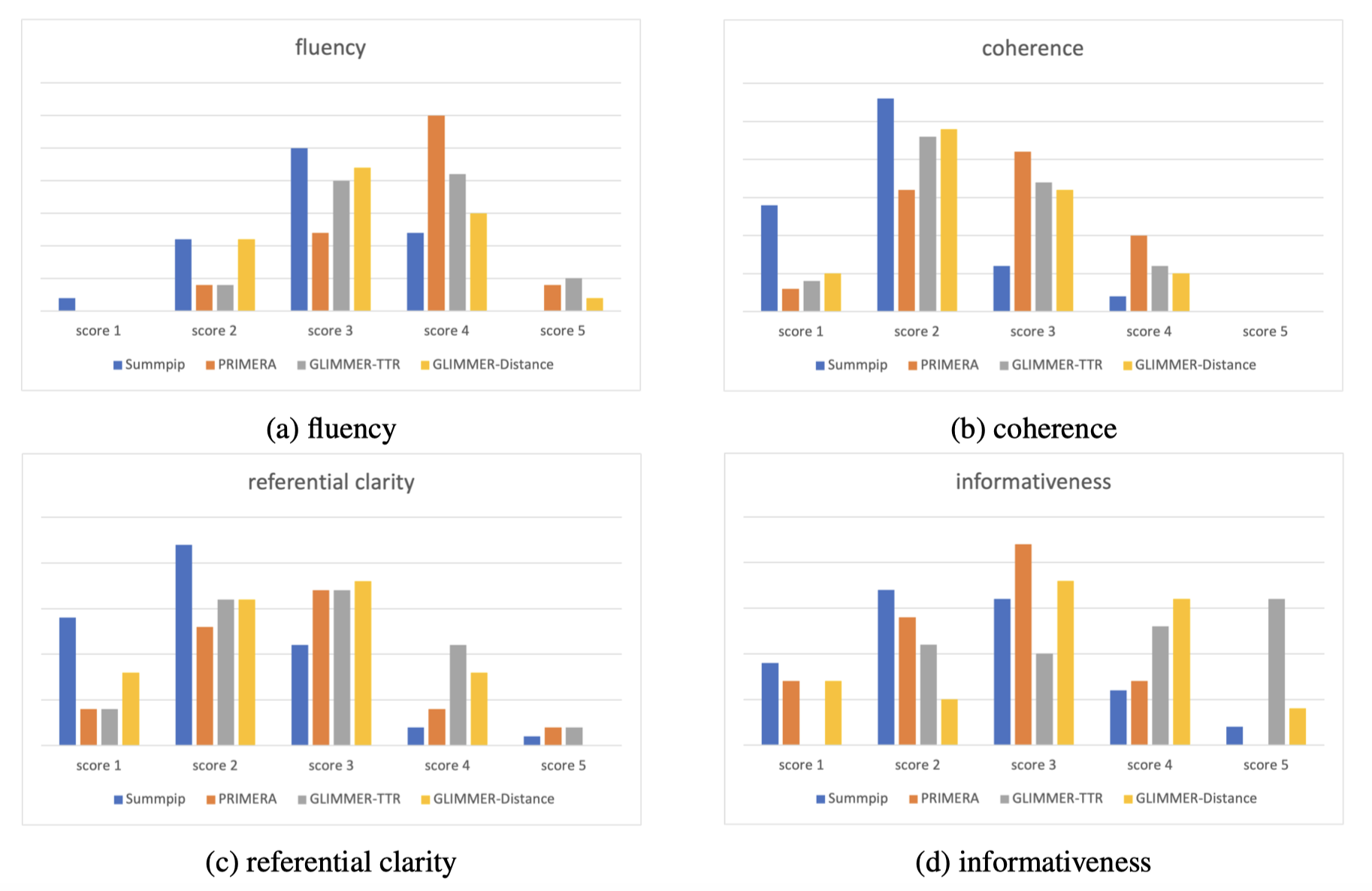}
    \caption{Distributions of human evaluation scores.}
    \label{human-distribution}
  \end{figure*}

  Figure~\ref{human-distribution} shows the distributions of human evaluation scores. Regarding fluency, the peak score for PRIMERA is 4, while scores 3 and 4 are both peaks for GLIMMER-TTR. Despite PRIMERA being a pre-trained language model, our unsupervised approach still has room for improvement in readability. However, in terms of informativeness, the peak score for GLIMMER-TTR is 5, whereas for PRIMERA it is 3, indicating that our approach is more consistent with the original text.
  
  \subsection{Agreement Analysis}
  \label{agreement}
  We compute Kendall's coefficient of concordance (Kendall-W) to measure inter-annotator agreement. Kendall-W assesses the agreement level among more than two annotators when scores are in rank order. According to Table \ref{agreement_table}, annotators achieved moderate agreement on fluency, coherence, and referential clarity, and substantial agreement on informativeness.

  The main inconsistency among the three annotators is that non-native English speakers have lower tolerance for minor grammar errors compared to the native English speaker. Additionally, some annotators perceive continuously repeated sentences as lacking fluency rather than incoherence. Nevertheless, we consider the agreement level among annotators to be acceptable, and the results of human evaluation are deemed reliable.

  \begin{table}
    \caption{\label{agreement_table}
    Agreement analysis.
    }
    \centering
    \begin{tabular}{lcc}
    \toprule
    indicator & Kendall-W & p-value \\
    \midrule
    fluency & 0.43 & .003 \\
    coherence & 0.51 & .000 \\
    referential clarity & 0.52 & .000 \\
    informativeness & 0.62 & .000 \\
    \bottomrule
    \end{tabular}
    \end{table}
  
\section{Evaluation by UniEval}
  \label{sec:unieval}
  UniEval was trained in the form of Boolean Question Answering, allowing it to evaluate summaries from multiple dimensions by providing different questions. To a certain degree, it correlates with human evaluation. Fluency, coherence, and consistency scores evaluated by UniEval are shown in Table~\ref{result-unieval}. Among these metrics, fluency and coherence assess linguistic quality, indicating the readability of summaries, while consistency reflects the factual alignment between summaries and source documents.

  Table~\ref{result-unieval} demonstrates that GLIMMER outperforms Summpip across all metrics. Despite GLIMMER's linguistic quality being inferior to that of fully fine-tuned PRIMERA, it significantly excels in consistency compared to PRIMERA. It's worth noting that as a neural network evaluation framework, UniEval may provide inappropriate scores in some cases. For instance, UniEval struggles to properly evaluate the performance of zero-shot PRIMERA. Although we have confirmed that zero-shot PRIMERA sometimes generates continuously repeated sentences, UniEval finds it challenging to identify such instances and may still provide very high fluency and coherence scores, even exceeding manually written reference summaries. Therefore, we omit the results for zero-shot PRIMERA, and we only consider the use of UniEval as a supplementary experiment.

  \begin{table*}
    \caption{\label{result-unieval}
    Evaluation results by UniEval on Multi-News. Summpip and fully fine-tuned PRIMERA are used as baselines.
    }
    \centering
    \begin{tabular}{lccc}
    \toprule
    model & fluency & coherence & consistency\\
    \midrule
    Summpip & 0.506 & 0.264 & 0.794\\
    PRIMERA & 0.591 & 0.467 &0.360\\
    GLIMMER-TTR & 0.549 & 0.351 & 0.818\\
    GLIMMER-Distance & 0.535 & 0.370 & 0.809\\
    \bottomrule
    \end{tabular}
    \end{table*}

\section{Estimation of D-Value}
    \label{sec:D}
    We use LexicalRichness\footnote{\url{https://github.com/LSYS/LexicalRichness}} to estimate D through the following steps:
    \\
    1. Randomly select 35 words from the input (a document set to be summarized), and calculate TTR value of this 35-word sample. Repeat the process 100 times and calculate average TTR.
    \\
    2. Repeat step 1 for samples of 36 words, 37 words, up to 50 words. This allows us to plot a TTR curve with respect to sample length.
    \\
    3. Use Formula~(\ref{ttr:estimate}) to identify the most appropriate value of D that best fits the curve obtained in step 2.
    \\
    4. Repeat steps 1 to 3 three times and calculate the average value of D.

\section{Detials of Comparison with ChatGPT}
    \label{sec:chatgpt}
    \subsection{Deployment Details}
    \label{detail-chatgpt}
    We utilize OpenAI API\footnote{\url{https://platform.openai.com}} to access ChatGPT. Apart from $model$ and $messages$, hyperparameters of API requests are consistent with official use case.\footnote{\url{https://platform.openai.com/docs/api-reference/chat/create}} We set $model$ to either \texttt{gpt-3.5-turbo} or \texttt{text-davinci-003}. For $messages$, we use "\textit{please summarize the following content in less than n words:}" as prompt, where n is set to 256, 128, 128 for Multi-news, Multi-XScience, DUC-2004, respectively. During our experiments, the API's response times varied widely due to network conditions and the OpenAI server, making it challenging to draw definitive conclusions on response times.
    \subsection{Human Evaluation of ChatGPT}
    \label{human-eval-chatgpt}
\begin{figure*}[htbp]
      \centering
      \includegraphics[width=0.7\linewidth]{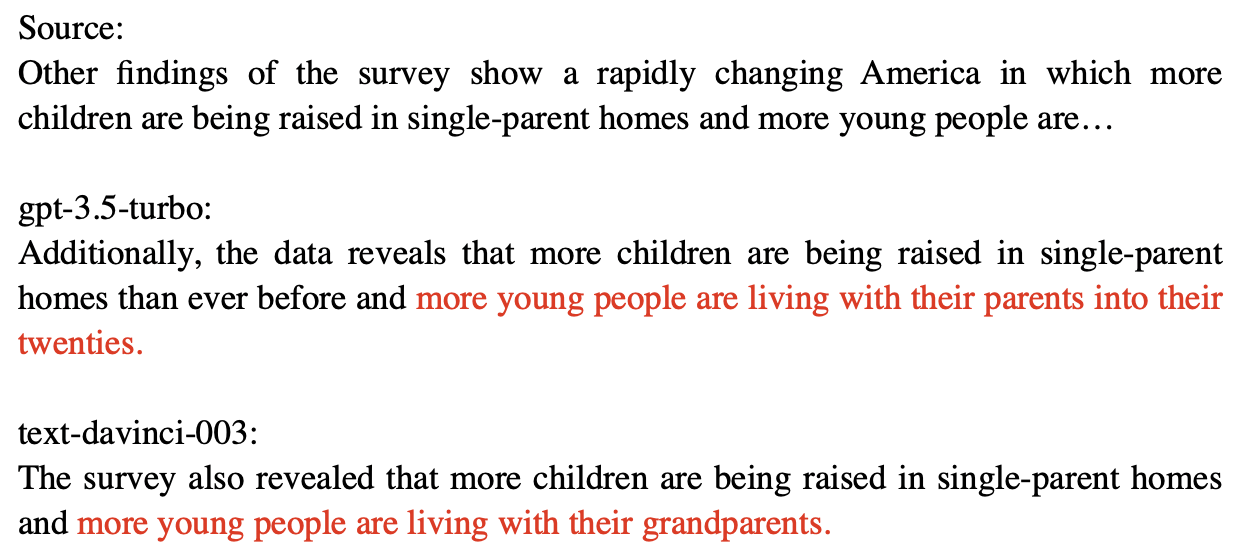}
      \caption{Hallucinations in ChatGPT.}
      \label{case-chatgpt}
    \end{figure*}

    We conduct human evaluation based on three readability indicators. Table \ref{chatgpt-human} presents human evaluation results of GLIMMER-TTR and two ChatGPT models on the first 50 samples of Multi-News. Despite their generally high readability, ChatGPT models may exhibit hallucinations. Some hallucinations are easily noticeable, for instance, text-davinci-003 might produce ungrammatical and inaccurate sentences from the outset. Other hallucinations are more subtle and can appear anywhere in summaries without obvious grammatical errors. Examples of such cases are shown in Figure~\ref{case-chatgpt}, where both gpt-3.5-turbo and text-davinci-003 generate inaccurate content, highlighted in red.

    \begin{table*}[htbp]
        \caption{\label{chatgpt-human}
        Human evaluation results of ChatGPT.
        }
      \centering
      \begin{tabular}{lccc}
      \toprule
      model & fluency & coherence & referential clarity\\
      \midrule
      GLIMMER-TTR & 3.85 & 3.77 & 3.58 \\
      gpt-3.5-turbo & \textbf{4.31} & \textbf{4.31} & \textbf{4.35} \\
      text-davinci-003 & 4.16 & 4.07 & 4.05 \\
      \bottomrule
      \end{tabular}
      \end{table*}

\section{Clustering Visualization}
    \subsection{TTR-based Method}
    \label{sec:pca}
\begin{figure*}[htbp]
        \centering
        \includegraphics[width=0.75\linewidth]{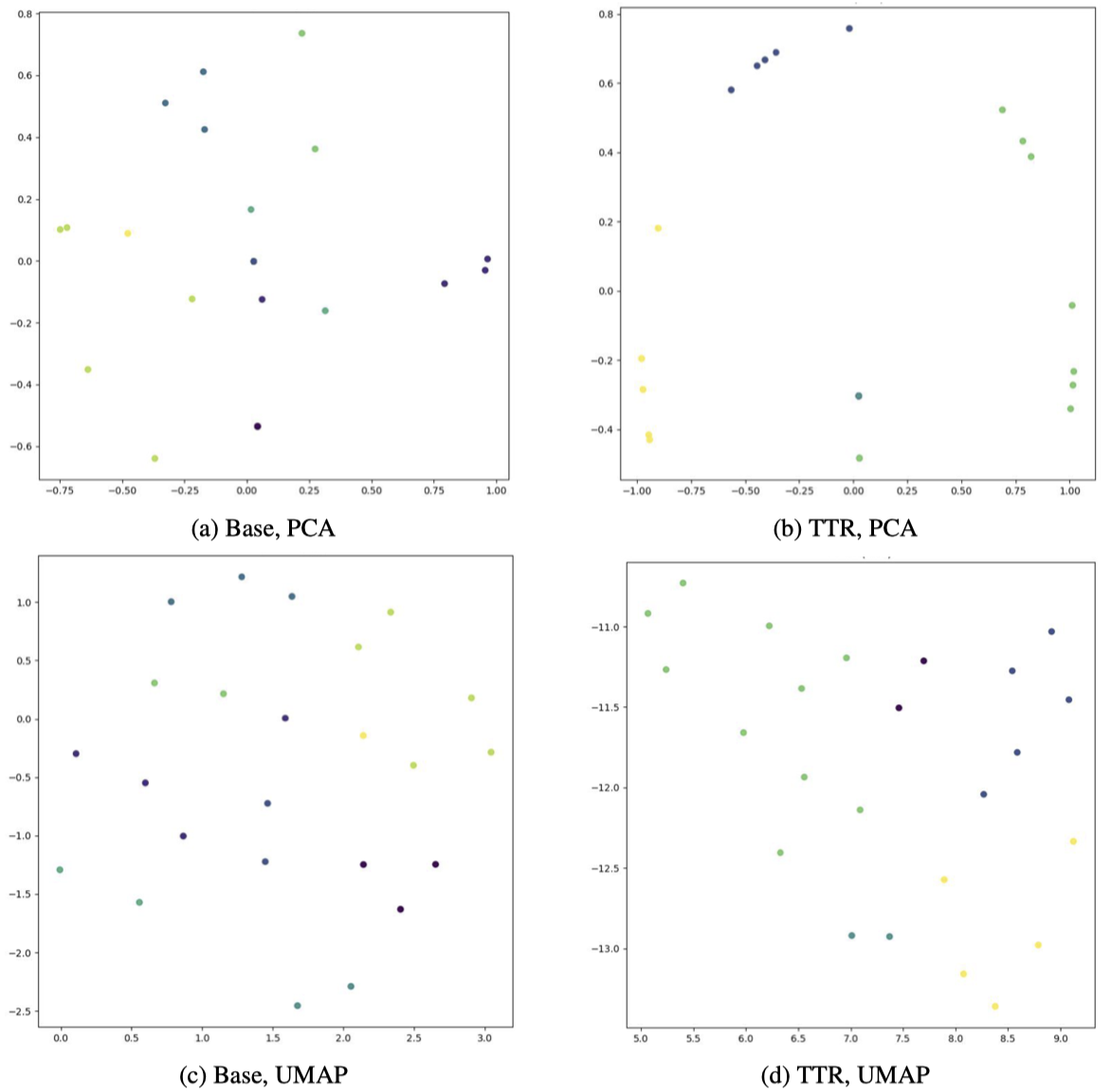}
        \caption{Visualization of clustering by the base model and TTR-based method, using PCA and UMAP to reduce dimension.}
        \label{fig-pca}
      \end{figure*}

    See Figure~\ref{fig-pca}.
    \subsection{Distance-based Method}
    \label{sec:cut}
\begin{figure*}[htbp]
        \centering
        \includegraphics[width=0.71\linewidth]{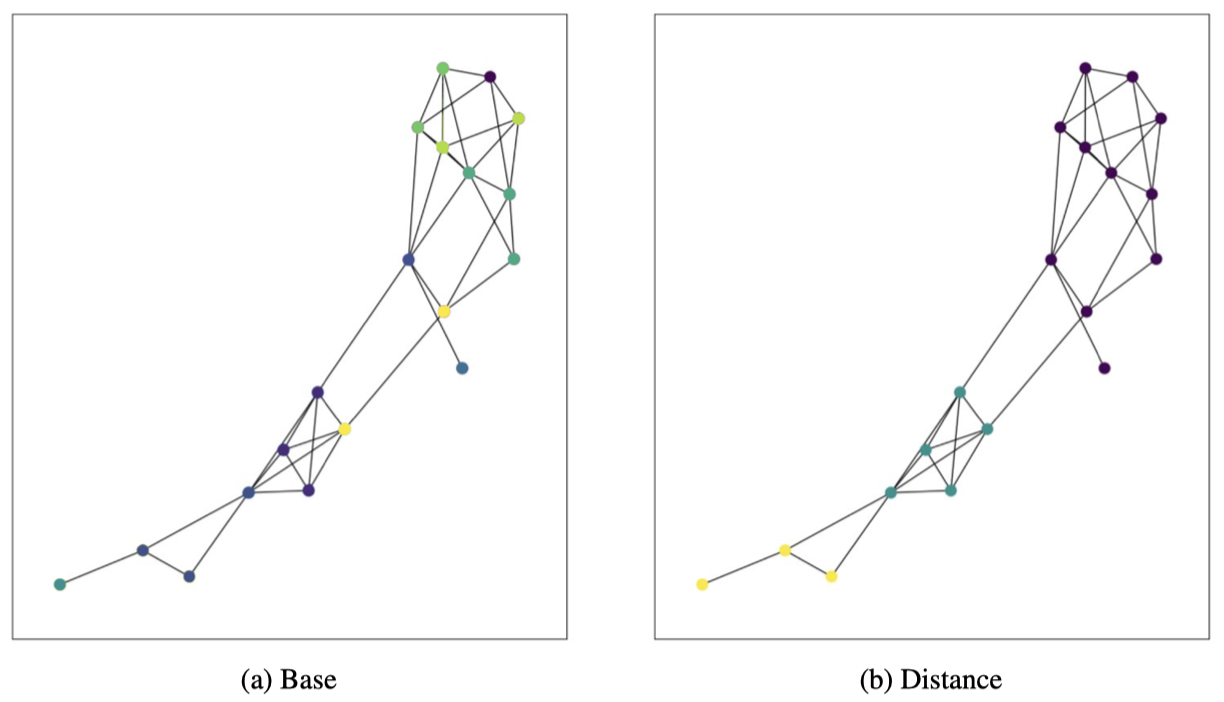}
        \caption{Visualization of graph cut by the base model and distance-based method.}
        \label{fig-cut}
    \end{figure*}

    See Figure~\ref{fig-cut}.

\section{Neural Cluster Summarization Models}
    \label{sec:compression}
    Four neural models are employed to summarize semantic clusters:
    \\
    \textbf{newsroom-L11} A model trained by \citet{ghalandari2022efficient} using Newsroom \citep{grusky2018newsroom} and to predict summary sentences of 11 tokens.
    \\
    \textbf{newsroom-P75} Similar with newsroom-L11 but trained to reduce clusters to 75\% of their original length. 
    \\
    \textbf{multi-news-P40} Trained by us on Multi-News to reduce clusters to 40
    \\
    \textbf{multi-news-Gamma} Unlike the previous three models that use a Gaussian distribution to control compression length, lengths of semantic clusters identified by GLIMMER follows a Gamma distribution. Therefore, we modified the reward function for length control, setting $\alpha$ to 2 and $\beta$ to 15.

\section{Software and Licenses}
    Our code will be released and licensed under Apache License 2.0. The framework dependencies include:
    \begin{itemize}
      \item scikit-learn,\footnote{\url{https://github.com/scikit-learn/scikit-learn/blob/main/COPYING}} BSD 3-Clause
      \item PyTorch,\footnote{\url{https://github.com/pytorch/pytorch/blob/main/LICENSE}} Misc
      \item NLTK,\footnote{\url{https://github.com/nltk/nltk/blob/develop/LICENSE.txt}} Apache 2.0
      \item LexicalRichness,\footnote{\url{https://github.com/LSYS/LexicalRichness/blob/master/LICENSE}} MIT
      \item SciPy,\footnote{\url{https://github.com/scipy/scipy/blob/main/LICENSE.txt}} BSD 3-Clause
      \item NetworkX,\footnote{\url{https://github.com/networkx/networkx/blob/main/LICENSE.txt}} BSD 3-Clause
      \item spaCy,\footnote{\url{https://github.com/explosion/spaCy/blob/master/LICENSE}} MIT
      \item Gensim,\footnote{\url{https://github.com/RaRe-Technologies/gensim/blob/develop/COPYING}} LGPL 2.1
      \item Gensim-data,\footnote{\url{https://github.com/RaRe-Technologies/gensim-data/blob/master/LICENSE}} LGPL 2.1
      \item ROUGE Metric,\footnote{\url{https://github.com/li-plus/rouge-metric/blob/master/LICENSE}} MIT
      \item Centroid,\footnote{\url{https://github.com/gaetangate/text-summarizer/blob/master/LICENSE}} GPL 3.0
      \item LexRank,\footnote{\url{https://github.com/crabcamp/lexrank/blob/dev/LICENSE.txt}} MIT
      \item PRIMERA,\footnote{\url{https://github.com/allenai/PRIMER/blob/main/LICENSE}} Apache 2.0
      \item UniEval,\footnote{\url{https://github.com/maszhongming/UniEval/blob/main/LICENSE}} MIT
      \item Multi-XScience,\footnote{\url{https://github.com/yaolu/Multi-XScience/blob/master/LICENSE}} MIT
      \item Multi-News,\footnote{\url{https://github.com/Alex-Fabbri/Multi-News/blob/master/LICENSE.txt}} Misc
      \item NumPy,\footnote{\url{https://github.com/numpy/numpy/blob/main/LICENSE.txt}} BSD 3-Clause
      \item OrderedSet,\footnote{\url{https://github.com/Weebly/OrderedSet/blob/master/LICENSE}} MIT
      \item matplotlib,\footnote{\url{https://github.com/matplotlib/matplotlib/blob/main/LICENSE/LICENSE}} Misc
    \end{itemize}

\end{document}